\documentclass[letterpaper, 10 pt, conference]{ieeeconf}  

\IEEEoverridecommandlockouts                              
                                                          
\usepackage{graphics} 
\usepackage{epsfig} 
\usepackage{mathptmx} 
\usepackage{times} 
\usepackage{amsmath} 
\usepackage{amssymb}  
\usepackage{algorithm}
\usepackage{algorithmic}
\usepackage{booktabs} 

\title{\LARGE \bf
Multi-agent Traffic Prediction via Denoised Endpoint Distribution
}

\author{Yao Liu$^{1}$, Ruoyu Wang$^{2}$, Yuanjiang Cao$^{1}$, Quan Z. Sheng$^{1}$, and Lina Yao$^{3}$
\thanks{Corresponding author: Yao Liu }
\thanks{$^{1}$Yao Liu, Yuanjiang Cao and Quan Z. Sheng are with School of Computing, Macquarie University, Sydney, Australia \tt\small y.liu@mq.edu.au; yuanjiang.cao@mq.edu.au; michael.sheng@mq.edu.au}%
\thanks{$^{2}$Ruoyu Wang is with School of Computer Science and Engineering, University of New South Wales, Sydney, Australia \tt\small ruoyu.wang5@unsw.edu.au}%
\thanks{$^{3}$Lina Yao is with Data 61, CSIRO \& School of Computer Science and Engineering, University of New South Wales, Sydney, Australia \tt\small lina.yao@unsw.edu.au}%
}

\begin{document}

\maketitle
\thispagestyle{empty}
\pagestyle{empty}

\begin{abstract}

The exploration of high-speed movement by robots or road traffic agents is crucial for autonomous driving and navigation. 
Trajectory prediction at high speeds requires considering historical features and interactions with surrounding entities, a complexity not as pronounced in lower-speed environments.
Prior methods have assessed the spatio-temporal dynamics of agents but often neglected intrinsic intent and uncertainty, thereby limiting their effectiveness.
We present the Denoised Endpoint Distribution model for trajectory prediction, which distinctively models agents' spatio-temporal features alongside their intrinsic intentions and uncertainties.
By employing Diffusion and Transformer models to focus on agent endpoints rather than entire trajectories, our approach significantly reduces model complexity and enhances performance through endpoint information.
Our experiments on open datasets, coupled with comparison and ablation studies, demonstrate our model's efficacy and the importance of its components. 
This approach advances trajectory prediction in high-speed scenarios and lays groundwork for future developments.

\end{abstract}


\section{Introduction}

Robot movement in low-speed, restricted scenarios can be autonomously navigated by analyzing the surrounding environment. 
However, in high-speed, open spaces, the robot must evaluate the trajectories and future intentions of surrounding objects to make informed judgments. 
Particularly in the rapidly evolving world of autonomous driving, accurately predicting vehicle movement trajectories is crucial for intelligent agents. 
This capability enables them to avoid collision risks without resorting to overly conservative maneuvers~\cite{Trajectron++, Trafficprecict}.

Early approaches to trajectory prediction primarily utilized traditional machine learning methods, such as Bayesian learning, Hidden Markov Models, Support Vector Machines, and Gaussian Processes. 
However, these methods often necessitate manual feature extraction and tend to be less robust, making it challenging to achieve satisfactory performance~\cite{Prediction_Gaussian, social_force}.

Recent advancements in deep learning have led to the perception of trajectory prediction as a quintessential sequence prediction problem. 
Recurrent Neural Networks (RNNs), particularly Long Short-Term Memory (LSTM) networks~\cite{Social_LSTM} and Gated Recurrent Units (GRU)~\cite{gru-compare}, are extensively employed. 
Concurrently, Convolutional Neural Networks (CNNs) processing raster map data and Graph Neural Networks (GNNs) addressing interactions have seen widespread use~\cite{Trajectron++}. 
Transformer-based methods, which have significantly impacted Natural Language Processing (NLP), are being successfully adapted for trajectory prediction~\cite{mmTransformer}. 
Owing to their multi-head attention mechanism and stacking structure, Transformers excel in managing complex and lengthy temporal relationships more effectively than RNNs.
Additionally, trajectory prediction has been approached as a generative task, utilizing Generative Adversarial Networks (GAN) and Variational Autoencoders (VAE)~\cite{Social_GAN, SoPhie}. 
Moreover, the introduction of diffusion models~\cite{mid} to trajectory prediction marks a significant development in the field.


Road traffic trajectory prediction, despite extensive study, continues to confront some challenges. 
The first challenge involves observable data. 
It is a well-established fact that traffic agents on the road are expected to adhere to corresponding road traffic rules, signs, lanes, maps, and so on. 
However, accurately obtaining this data in real-time proves difficult in practical applications. 
Consequently, predicting future trajectories based solely on the agent's historical trajectories becomes crucial. 
This approach should be applicable across various autonomous driving scenarios, irrespective of the availability of additional auxiliary information in the agent's environment. 
The second challenge pertains to complex spatio-temporal relationships. 
Agents must consider historical trajectories in their movement, which encompass critical time points where decisions significantly impact future trajectories. 
Simultaneously, agents must account for the presence of neighboring agents, choosing to either follow or avoid them, thereby adopting different maneuvers. 
The third challenge is the agent's own intent. 
The final destination or maneuver intent of a transport agent, especially when human-driven, is often self-determined. 
That is, historical trajectories of transport agents do not necessarily predict similar future trajectories; a human driver's final movement is dictated by intrinsic intentions. 
This introduces an inherent uncertainty~\cite{Intention-Aware}.

\begin{figure}[htbp]
  \centering
  \includegraphics[width=0.9\linewidth]{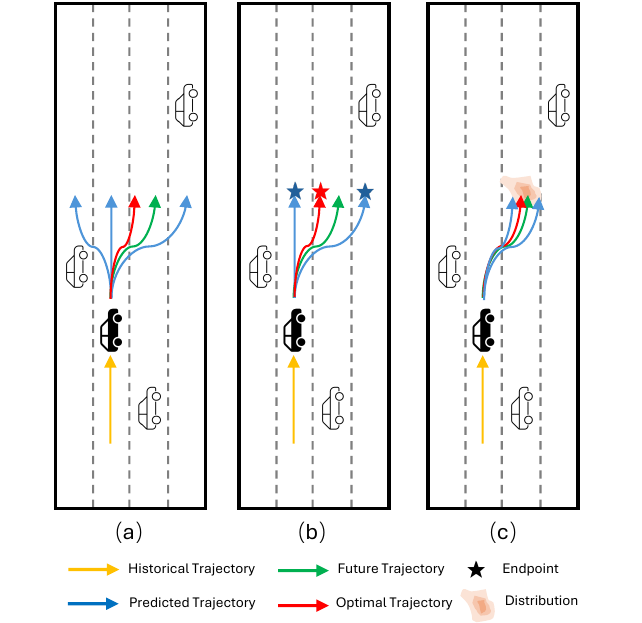}
  \caption{Trajectory prediction scenarios. (a) Sequence Prediction, (b) Endpoint-based Prediction, (c) Denoised Endpoint Distribution-based Prediction (ours).}
  \label{figure-example}
\end{figure}

In light of these challenges, we introduce a Denoised Endpoint Distribution model for multi-agent trajectory prediction, as illustrated in Figure~\ref{figure-example}. 
Our model exclusively relies on the agent's historical trajectory as its sole input, deliberately eschewing additional auxiliary information such as maps and lanes. 
This design choice aims to minimize the model's complexity and computational demands, thereby offering real-time and practically meaningful solutions for robotics and autonomous driving domains. 
Furthermore, our model is capable of simultaneously predicting the trajectories of all agents present in the scene, rather than focusing on a single agent. 
This approach treats all agents equally, considering their interactions to be mutual. Unlike previous methodologies~\cite{cslstm-pre, gru-compare} that differentiate between target and neighboring agents—whereby they can predict only one target agent at a time, using neighbor agents solely for interaction information—our model integrates and processes all agents collectively.

Specifically, we assert that an agent's trajectory is shaped by spatio-temporal information from historical trajectories and neighbor interactions, while also inherently encompassing motion uncertainty. 
To this end, we first utilize a spatio-temporal graph network to gather historical and interaction features as foundational guidance. 
Following this, a diffusion model is applied to craft the distribution of endpoints for each trajectory. 
This phase leverages spatio-temporal guidance to produce endpoint distributions via the denoising process of inverse diffusion, representing the density distribution of potential endpoints influenced by historical actions and neighbor interactions.
Subsequently, we predict the trajectory's endpoints using the Transformer model, focusing solely on the agent's historical trajectory. 
This prediction, uninfluenced by neighbor interactions, aims to reflect the agent's intrinsic intent.
Given the inherent motion uncertainty, the endpoints predicted in this step might vary. 
Therefore, we employ the previously established distribution to filter and pinpoint the endpoints that most closely adhere to the distribution, designating these as the trajectory's predicted endpoints.
In the final step, we predict the entire trajectory, integrating the initial guidance with the endpoint information. 
This method enables a comprehensive trajectory prediction, factoring in both the influence of spatio-temporal dynamics and the agent's individual intent.

Our model presents several advantages. 
First, we simplify the perceived complexity of diffusion models by focusing on modeling trajectory endpoints rather than the entire trajectory during the initial diffusion step, which significantly reduces model complexity. 
Moreover, our endpoint distribution and prediction modules specifically address the spatial-temporal dimensions and intrinsic intent of trajectories, thereby enhancing prediction accuracy. 
The concept of intrinsic intent accounts for the uncertainty associated with divergent future trajectories emanating from identical historical paths. 
Concurrently, the endpoint distribution, anchored in spatial-temporal characteristics, finely tunes endpoint predictions. 
Additionally, employing predicted endpoint information streamlines the trajectory prediction process, enabling a more efficient model in the final prediction phase. 
This strategy results in a lightweight model that minimizes dimensionality, optimizing both efficiency and accuracy in the context of multi-agent trajectory forecasting.

Our main contributions are as follows:
\begin{itemize}
\item Our Denoised Endpoint Distribution model simultaneously predicts the future trajectories of all multi-agents in the scenario using only the historical trajectories of the agents, effectively reducing the complexity and increasing the practicality of the model.
\item Our model explicitly captures the spatio-temporal features and intrinsic intent of trajectories through the endpoint distribution module and endpoint prediction module, which effectively improves the performance of the model.
\item Our model has been experimented on realistic datasets with state-of-the-art results, and the corresponding comparison and ablation experiments highlight the strengths of the model and the contribution of the module.

\end{itemize}

\section{Related Works}

\subsection{Sequential Models}

Trajectory prediction, commonly framed as a sequence problem, has seen significant application from the inception of deep learning, particularly with RNN models, including LSTM and GRU. 
Social-LSTM~\cite{Social_LSTM} marked a pioneering effort by introducing a pooling mechanism to aggregate spatial feature information surrounding an agent, employing LSTM to generate trajectory predictions. 
Trafficpredict~\cite{Trafficprecict} models social interactions between agents using soft attention, leveraging LSTM to capture movement similarities within the same category for predictions. 
However, these approaches, which depend on LSTM's individual vectors to chronicle history, may stumble when handling complex time-dependent sequential data.

Transformers~\cite{Transformer_Trajectory} have shown promising outcomes by directly applying self-attention mechanisms for pedestrian trajectory prediction, sidestepping the need for explicit consideration of social interactions or scene contexts.
mmTransformer~\cite{mmTransformer} integrates stacked transformers as a central component for trajectory prediction, amalgamating historical trajectories, social interactions, and map data. 
The cornerstone of Transformers' success lies in their self-attention mechanism, enabling comprehensive sequence consideration. 
This feature positions them ideally for modeling sequential data and predicting trajectories.

\subsection{Generative Models}

Some methodologies conceptualize trajectory prediction as a generative task. 
Social-GAN~\cite{Social_GAN} integrates a novel pooling mechanism with a GAN model to predict trajectories. 
DESIRE~\cite{DESIRE} employs a multimodal sampling strategy, utilizing a Conditional Variational Auto-Encoder (CVAE) within a sampling-based Inverse Optimal Control (IOC) framework to generate multiple future trajectories. 
SoPhie~\cite{SoPhie} extracts scene information using a CNN, followed by the application of a bidirectional attention mechanism to distill social features. 

Recently, diffusion modeling, a generative approach inspired by nonequilibrium thermodynamics, has gained notable interest. 
MID~\cite{mid} is among the first to apply diffusion modeling to trajectory prediction, treating the prediction process as a denoising task that transforms Gaussian noise into discernible paths. 
However, the substantial computational cost often associated with diffusion modeling poses limitations~\cite{leap}.

\subsection{Multimodal Prediction}

Social-GAN~\cite{Social_GAN} introduces the notion of multimodal prediction, advocating that the domain of trajectory prediction should encompass the generation of multiple viable trajectories based on principles, with the anticipation that future paths will conform to one of these trajectories. 
TPNet~\cite{TPNet} innovates with a two-stage motion prediction framework that initially creates a pool of potential future trajectories, subsequently refining the final prediction to align with physical constraints. 
Maneuver~\cite{Maneuver} incorporates the consideration of multiple maneuvering intentions, thus enabling multimodal predictions. 
Multimodal prediction has become a standard practice in trajectory prediction, offering a strategy to partially counterbalance the inherent uncertainty of motion.

\section{Methodology}

\begin{figure*}[htbp]
  \centering
  \includegraphics[width=0.9\linewidth]{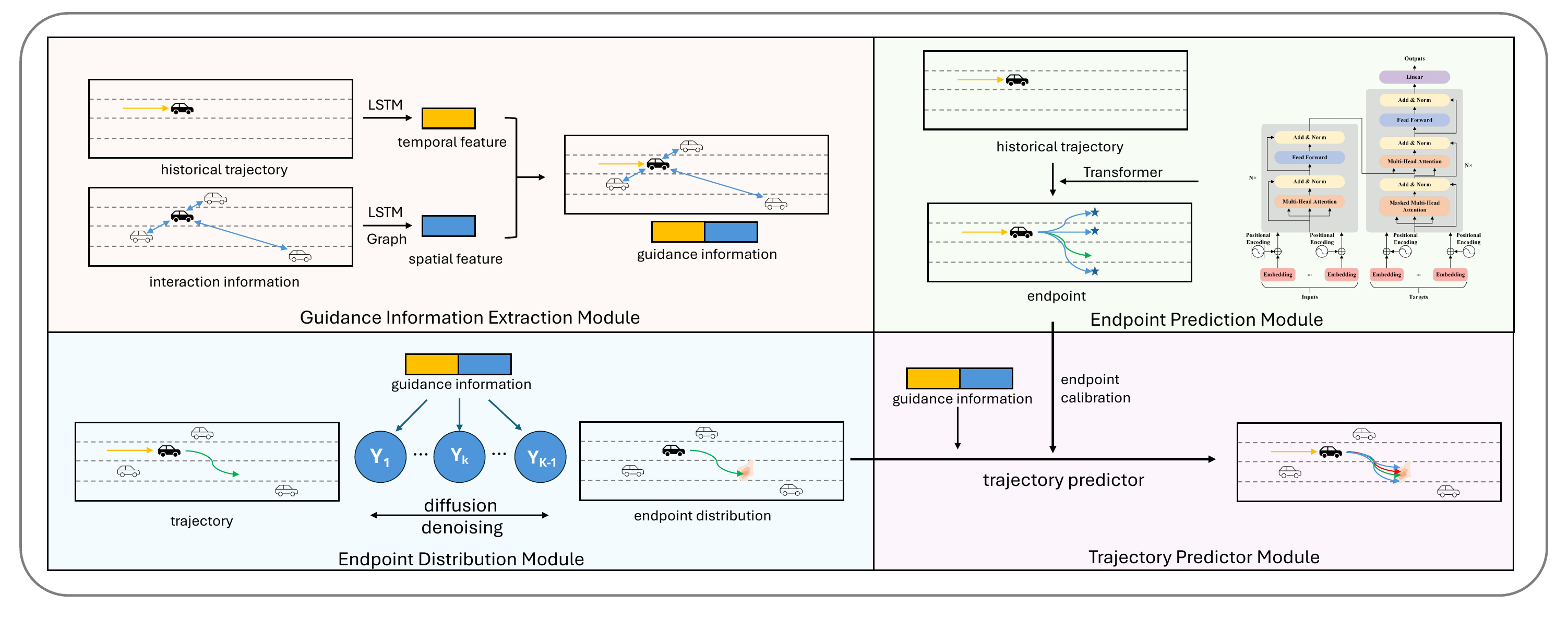}
  \caption{The overview of our model. Our model consists of four main modules, i.e. Guidance Information Extraction, Endpoint Distribution, Endpoint Prediction and Trajectory Prediction.}
  \label{figure-overview}
\end{figure*}

\subsection{Problem Definition}

Our model forecasts future trajectories based on historical trajectories, which are represented as a series of coordinate points arranged in a temporal sequence.

$x_n^t \in \mathbb{R}^2$ denotes the coordinates of the position of the agent $n$ in the current scene at time $t$, $\mathbf{x}_n= \{x_n^1, \dots, x_n^{t_{\text{past}}}\}$ denotes the historical trajectory of agent $n$, and $\mathbf{X} = \{\mathbf{x}_1, \dots, \mathbf{x}_N\}$ denotes the set of multi-agent historical trajectories. Note that $n$ can be different at different moments, i.e., agents can come in and out of the current scene at will, and the number of agents in each scene is not fixed. Correspondingly, $\mathbf{y}_n = \{y_n^{t_{\text{past}+1}}, \dots, y_n^{t_{\text{past}}+t_{\text{future}}}\}$ denotes the future trajectory of agent $n$, and $\mathbf{Y} = \{\mathbf{y}_1, \dots, \mathbf{y}_N\}$ denotes the set of multi-agent future trajectories.
The final prediction of our model is a multimodal trajectory, i.e., multiple plausible trajectories with a view to one of them following the true future trajectory $\hat{\mathbf{Y}} \sim {\mathbf{Y}}$.

\subsection{Model Overview}

Our model contains four main modules, which are Guidance Information Extraction module (GE), Endpoint Distribution module (ED), Endpoint Prediction module (EP), and Trajectory Predictor module (TP), as shown in Figure~\ref{figure-overview}.

Briefly, in the GE module, we model the historical trajectories of the agents as temporal features by LSTM model, then model the neighbor information as spatial features by graph aggregation function, and finally get the guidance information of the agents as spatio-temporal features by connection. Note that in this processing, all the multi-agents are processed at the same time and they are neighbors to each other without targeting a particular target agent.
In Figure~\ref{figure-overview}, only one agent is focused on for ease of representation, but the model is handling all the agents simultaneously.

In the ED module, we model the endpoint distribution of the trajectory using the generative model. Specifically, we adopt a diffusion model that uses the spatio-temporal features of the trajectories as a guide for modeling, and gradually restores the endpoint distributions of the trajectories through denoising in the reverse diffusion process. 
The estimated endpoint distribution contains the spatio-temporal features of the trajectories, indicating that the probability density of the endpoints is affected by the historical trajectories of the agents and their neighborhood features.

In the EP module, we use the Transformer to model the historical features of agents in order to predict the endpoints of trajectories. The endpoints of the trajectories predicted in this way are only influenced by the agent itself, incorporating the agent's intrinsic intent and uncertainty, and thus may be able to become divergent themselves, indicating the multimodal nature that the same historical trajectories may not follow the same future trajectories. 

We therefore need to filter the endpoints that best fit the current situation by the distribution of endpoints, a process we refer to as calibration, i.e., selecting an endpoint that is most likely to occur in the current scenario among the multiple uncertain intentions of the agent itself.
Finally, we predict the whole trajectory by using the predicted endpoints as well as the spatio-temporal guidance information by the TP module.

\subsection{Guidance Information Extraction}

Here we first describe our processing of the data and the extraction of spatio-temporal features as guidance information to be used in subsequent modules.

For multi-agent historical trajectories, we model them as spatio-temporal graphs, where each agent in the temporal dimension has the location of a historical point in time, and each agent in the spatial dimension has neighbor agents. 
For the input data, we compute the velocity and acceleration of agent by coordinates with time intervals to enrich the model feature dimensions. Note that velocity and acceleration are computed without additional auxiliary information.

First we model the historical trajectory of the agent through LSTM to obtain the temporal feature ($temp_n$) of the current agent. This high-dimensional temporal feature represents the current and historical state of the agent, which is not affected by the interaction information of the neighbors, and thus contains the agent's intrinsic intention as well as uncertainty.
Earlier methods~\cite{Transformer_Trajectory} that did not consider interaction information used this feature for prediction, usually leading to divergent multimodal prediction results.
This divergence indicates that the agent is affected by its intrinsic intent and uncertainty resulting in the same historical trajectory not necessarily following the same future.

We then aggregate the neighbor features around the agent through the graph network to obtain the interaction information as spatial information ($spat_n$), which contains the current and historical states of the neighbors.
The advantage of using graph networks is that it is easier to aggregate a varying number of neighboring features, since the number of neighbors of an agent at each moment in time is not fixed.

Finally we connect the temporal and spatial features to become the guidance information ($gi_n$) for the spatio-temporal features.
Most current methods~\cite{Social_LSTM} make predictions with guidance information that contains the agent's spatio-temporal features; however, such methods submerge the agent's intrinsic intent and uncertainty within the spatio-temporal features,  leading to excessive displacement errors.

\begin{equation}
temp_n=LSTM(\mathbf{x_n})
\end{equation}
\begin{equation}
spat_n=Agg(LSTM(\mathbf{x_n}),~LSTM(\mathbf{X}))
\end{equation}
\begin{equation}
gi_n = Concat(temp_n,~spat_n)
\end{equation}

\subsection{Endpoint Distribution Module}

\begin{figure}[htbp]
  \centering
  \includegraphics[width=1\linewidth]{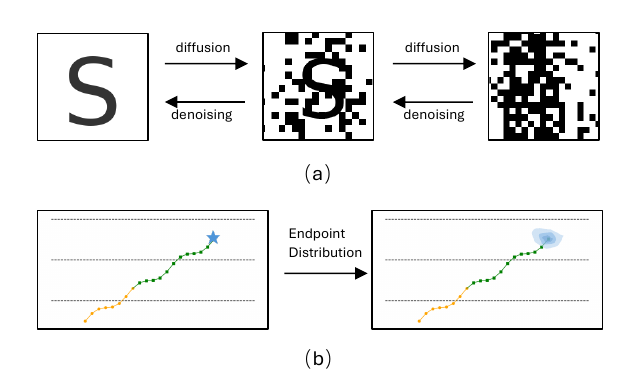}
  \caption{Diffusion and denoising model. (a) Diffusion and denoising process for image generation (b) Generation process for endpoint distribution.}
  \label{figure-diffusion}
\end{figure}

As shown in Figure~\ref{figure-diffusion}(a), the diffusion model is based on nonequilibrium thermodynamics, and the diffusion process is simulated by adding random noise to the image until it becomes Gaussian noise, after which the image is restored by gradually removing the noise from the Gaussian noise.

We use a diffusion model to generate endpoint distributions, as shown in Figure~\ref{figure-diffusion}(b). The goal is to obtain a distribution of future trajectory endpoints from the agent's guidance information, and this endpoint distribution represents the probability density of the positions that the agent may reach if it is influenced by the historical trajectory and neighbor interaction information.

We use the diffusion model to generate the endpoint distribution. 
Specifically, we model the diffusion process by continuously adding noise to the endpoint distribution to make it Gaussian. This process indicates that the agent endpoints can be uniformly distributed throughout the space when the uncertainty of the agent gradually increases to the limit. In the denoising process of reverse diffusion, we iteratively remove the noise starting from randomly sampled Gaussian noise to restore the distribution of endpoints. The process can be expressed as the uncertainty of the endpoints gradually decreases under the influence of the agent's guidance information to obtain a more centralized probability density.

$\mathbf{Y_0}$ is denoted as the endpoint distribution as the beginning of the diffusion process, and $\mathbf{Y_K}$ denotes the Gaussian noise after $K$ steps of diffusion, then the following relation is obtained:

\begin{equation}
q(\mathbf{Y_{1:K}}|\mathbf{Y_0}) = \prod_{K}^{k=1}q(\mathbf{Y_k}|\mathbf{Y_k-1})
\end{equation}
\begin{equation}
q(\mathbf{Y_{k}}|\mathbf{Y_{k-1}}) = \mathcal{N}(\mathbf{Y_{k}};~\sqrt{\alpha_k}\mathbf{Y_{k-1}},~(1-\alpha_k)\mathbf{I})) 
\label{eq-a1}
\end{equation}

\begin{equation}
q(\mathbf{Y_{k}}|\mathbf{Y_{0}}) = \mathcal{N}(\mathbf{Y_{k}};~\sqrt{\bar{\alpha}_k}\mathbf{Y_{0}},~(1-\bar{\alpha}_k)\mathbf{I}))
\label{eq-a2}
\end{equation}

In Equation~\ref{eq-a1, eq-a2}, $\alpha$ represents the weight coefficient that decreases gradually and uniformly with the diffusion step, indicating the rise in the weight of the noise added gradually to the endpoint distribution. 
By calculation, the diffusion features of adding $K$-step noise can be obtained directly from the initial true distribution, i.e., there is no need to add noise through sequential iterations, and the final diffusion features can be obtained directly from the initial endpoint distribution by calculation at one time.

In the reverse process, we start with Gaussian noise $\mathbf{Y_{K}}$ and gradually restore the distribution of endpoints through iterative denoising, which can be expressed as:

\begin{equation}
p_\theta(\mathbf{Y_{0:K}}|f)=p(\mathbf{Y_{K}})\prod_{K}^{k=1}p_\theta(\mathbf{Y_{k-1}}|\mathbf{Y_{k}},f)
\end{equation}
\begin{equation}
p_\theta(\mathbf{Y_{k-1}}|\mathbf{Y_{k}},f)=\mathcal{N}(\mathbf{Y_{k-1}};~\mu_\theta(\mathbf{Y_{k}},~k,~f);~(1-\alpha_k)\mathbf{I})
\end{equation}

The denoising process is limited by the fact that the Markov process requires iterative computation, which is the main reason for the high overhead of the diffusion model. However, our model only models the endpoint distribution of the trajectory rather than the overall trajectory thus saving computational resources.

However, optimizing the log-likelihood in an inverse process is extremely difficult, so optimization in the form of maximizing a variational lower bound is commonly used. By calculating the KL divergence of $q(\mathbf{Y_{k-1}}|\mathbf{Y_k},\mathbf{Y_0})$ and $p_\theta(\mathbf{Y_{k-1}}|\mathbf{Y_k},f)$ we can obtain:

\begin{equation}
q(\mathbf{Y_{k-1}}|\mathbf{Y_{k}},~\mathbf{Y_{0}}) = \\ \mathcal{N}(\mathbf{Y_{k-1}};~ \tilde{\mu}_k(\mathbf{Y_k},\mathbf{Y_0});~\tilde{\alpha}_k\mathbf{I}
\end{equation}

\begin{equation}
\tilde{\mu}_k(\mathbf{Y_k},\mathbf{Y_0}) = \frac{\sqrt{\bar{\alpha}_{k-1}}(1-\alpha_k)}{1-\bar{\alpha}}\mathbf{Y_0+\frac{\sqrt{\alpha}(1-\bar{\alpha}_{k-1})}{1-\bar{\alpha}_k}\mathbf{Y_k}}
\end{equation}

\begin{equation}
\tilde{\alpha}_k=\frac{1-\bar{\alpha}_{k-1}}{1-\bar{\alpha}_{k}}(1-\alpha)
\end{equation}

$\tilde{\alpha}_k$ can be computed at each $k$-step and $\tilde{\mu}_k(\mathbf{Y_k},\mathbf{Y_0})$ can be expressed as by reparameterization:

\begin{equation}
\mu_\theta(\mathbf{Y_k,~k,~f})=\frac{1}{\sqrt{\alpha_k}}(\mathbf{Y_k}-\frac{1-\alpha_k}{\sqrt{1-\bar{\alpha}_k}}\epsilon_\theta(\mathbf{Y_k,~k,~f})))
\end{equation}

The final simplified Loss is the MSE of the noise computed by the model $\theta$ with the noise added in.
The Endpoint Distribution Module enables us to generate an agent's endpoint distribution ($ed_n$) with guidance information ($gi_n$). Instead of using $ed_n$ directly for trajectory prediction, we use it to calibrate the endpoints for subsequent predictions. The reason for this is that $ed_n$ submerges the agent's intrinsic intent and uncertainty in the spatio-temporal features, leading to a limitation in prediction performance.

\subsection{Endpoint Prediction Module}

Transformer gradually surpasses RNN in sequence data processing, and its core lies in the use of multi-head self-attention mechanism instead of recursive structure. 
Transformer adopts multi-head self-attention mechanism to better deal with complex nonlinear temporal dependencies, whereas RNN can only deal with current and previous temporal embeddings. 
Through positional encoding and mask, Transformer does not need recursive computation, thus improving efficiency. 

Here we use Transformer to predict the location of endpoints based on the agent's historical trajectory. Transformer processes the agent's complex historical trajectory to capture features at key time points, and the predicted endpoints are only affected by the temporal features of the agent's own history and not by neighbor interactions. The predicted endpoint locations can be more divergent, which indicates that in multimodal trajectory prediction, the same historical trajectories do not necessarily follow the same future trajectories, as affected by the inherent intent and uncertainty of the agnet. When unaffected by neighbor interactions, the location of endpoints can go in many different directions based on the agent itself alone.

Specifically, Transformer's self-attention mechanism encodes three vectors independently through a linear mapping. 
They are the query vector ($\mathbf{Q}$), the key vector ($\mathbf{K}$), and the value vector ($\mathbf{V}$). The query vector and key vector compute the attention score and then the sum vector computes the attention value.

\begin{equation}
\mathbf{Q}=f_{Q}(\mathbf{Hist}),~\mathbf{K}=f_{K}(\mathbf{Hist}),~\mathbf{V}=f_{V}(\mathbf{Hist})
\end{equation}
\begin{equation}
Attention(\mathbf{Q},\mathbf{K},\mathbf{V})=softmax(\frac{\mathbf{Q}\cdot \mathbf{K}^T}{\sqrt{d_k}})\mathbf{V}
\end{equation}

The superscript $T$ denotes the transpose of the vector, $d_k$ denotes the dimension of each query, and $\frac{1}{\sqrt{d_k}}$ is used to scale the dot product to stabilize the gradient.

Transformer learns temporal dependencies over long periods of time by capturing self-attention at each time point, making it more suitable for trajectory prediction tasks than RNNs that use individual vectors to encode historical information. Meanwhile, Transformer encodes self-attention independently as queries, and keys and values help to learn complex historical trajectory temporal dependencies.

The performance of self-attention can be further improved by a subsequent multi-head mechanism that introduces multiple representation subspaces, and joint processing of these subspaces allows the model to combine several different hypotheses to focus on information features at different positions.

\begin{equation}
MultiHead(\mathbf{Q},\mathbf{K},\mathbf{V})=f(concat(head_1,\cdots,head_H))
\end{equation}
\begin{equation}
head_h=Attention_h(\mathbf{Q},\mathbf{K},\mathbf{V})
\end{equation}

In this module we predict the location of endpoints ($ep_n$), which can be more divergent according to the multimodal trajectory prediction paradigm.

\subsection{Trajectory Predictor Module}

In the endpoint distribution module we obtain endpoint distributions ($ed_n$) that are subject to information about the agent's spatio-temporal guidance, and in the endpoint prediction module we obtain multimodal endpoint locations ($ep_n$) that are only affected by the agent itself.
$ed_n$ denotes the statistically significant space of endpoint distributions that an agent may reach subject to historical and neighbor interaction information; $ep_n$ denotes the multimodal endpoint locations that an agent may reach subject to intrinsic intent and uncertainty.
We use $ed_n$ to calibrate $ep_n$ to denote the endpoints that are optimally likely to be reached by the agent under the influence of current historical information and neighbor interactions.
In this way, we show that independently modeling the distribution space where the agent is influenced by spatio-temporal guidance information, as well as the multiple possible endpoint locations that are only internally influenced. 
It does not submerge intrinsic intentions and uncertainties within the complex guidance information, but also ensures that the agent adequately perceives the influence of the guidance information.

Specifically, We consider the NLL scores of multiple $ep_n$ and $ed_n$, and choose the $ep_n$ that best matches $ed_n$ as the current agent's endpoint location.

\begin{equation}
ep_n=Select(NLL(multi(ep_n), ed_n))
\end{equation}

Finally, we use agnet's historical trajectory, guidance information, and endpoint location as inputs for agent's trajectory prediction through Transformer. Due to the endpoint location, Transformer does not have to use higher dimensional feature extraction, which can reduce the complexity of the model while the endpoint location also ensures the speed and performance of the model convergence.
\begin{equation}
\mathbf{Y'} = Transformer(\mathbf{X},~gi,~ep)
\end{equation}
\begin{equation}
 \mathbf{Y'} \sim \mathbf{\hat{Y}} = \mathcal{N}(\hat{\mu},~\hat{\sigma},~\hat{\rho })
\end{equation}
 
Since we follow multimodal prediction, $\mathbf{Y'}$ is a bivariate Gaussian distribution of the predicted trajectory $\mathbf{\hat{Y}}$, which will be sampled.
The bivariate Gaussian distribution contains five parameters, which are the mean $\mu \in \mathbb{R}^2 $, standard deviation $\sigma \in \mathbb{R}^2$, and correlation coefficient $\rho$.


\section{Experiments}

\subsection{Datasets and Settings}

We evaluate our model using the publicly available NGSIM dataset~\cite{i80, us101}, which comprises real-time vehicle trajectory data from two U.S. highways: US-101 and I-80. This dataset includes various traffic states—light, moderate, and congested—across sub-datasets, each lasting for 15 minutes and sampled at a frequency of 10Hz.

In alignment with established benchmark practices~\cite{PiP,han}, we implement 2x downsampling and partition the data into 8-second intervals. These intervals consist of the first 3 seconds (15 frames) representing historical trajectories and the subsequent 5 seconds (15 frames) denoting future trajectories. The dataset is divided into segments used for training (70\%), validation (20\%), and testing (10\%).

Our model's predictive accuracy is assessed using the Root Mean Square Error (RMSE), with evaluation checkpoints spanning from the first to the fifth second of the future trajectory.

\subsection{Experiments}

\subsubsection{Comparison Experiments}

We benchmark the performance of our model against several advanced models, with the comparative results presented in Table~\ref{table_comp}.

\begin{itemize}
\item Constant Velocity (CV): This approach predicts future positions based on constant velocity, utilizing the Kalman filter for estimation.

\item Vanilla LSTM (V-LSTM): This method employs a basic LSTM architecture for trajectory prediction, without incorporating surrounding context.

\item C-CGMM+VIM~\cite{cslstm-pre}: This technique leverages a maneuver-based variational Gaussian mixture model combined with a Markov random field module to account for vehicle interactions.

\item S-GAN~\cite{Social_GAN}: This model uses a Generative Adversarial Network (GAN) to generate future trajectories.

\item GAIL-GRU~\cite{gru-compare}: This approach integrates generative adversarial modeling with gated recurrent units for prediction, generating 20 trajectories and incorporating ground truth data from surrounding vehicles.

\item CS-LSTM~\cite{cslstm}: CS-LSTM aggregates surrounding interaction information through a social convolutional pool and employs LSTM for prediction. It includes the base model and a multimodal variant with an added classifier (CS-LSTM (M)).

\item GRIP~\cite{GRIP}: It utilizes graph structures to aggregate surrounding interaction information, coupled with LSTM for prediction.

\item PiP~\cite{PiP}: This model predicts trajectories by correlating historical data with planning insights.

\item HAN~\cite{han}: It uses a spatio-temporal attention module to extract interactive information from neighboring agents, capturing both historical and planning data as time-dependent factors for prediction.

\item PECNet~\cite{pcenet}: Pioneering in Endpoint-conditional prediction, it initially estimates pedestrian targets and then predicts trajectories based on these conditions. 
We reproduce its variants as vehicle trajectory prediction models.

\item MID~\cite{mid}: As an early attempt to incorporate diffusion modeling into trajectory prediction, MID approximates the reverse diffusion process to progressively diminish prediction uncertainty, yielding accurate trajectory predictions for each agent. 
We reproduce its variants as vehicle trajectory prediction models.
\end{itemize}

\begin{table}[htbp]
\caption{Comparison of prediction results on NGSIM. The value is RMSE in meters. Each model reports the prediction error from the 1st to the 5th second. The best results are indicated in bold, while the second-best results are underlined.}
\centering
\label{table_comp}
\resizebox{0.8\linewidth}{!}{
\begin{tabular}{l|ccccc}
\toprule
\begin{tabular}[c]{@{}l@{}}Prediction\\ Horizon(s)\end{tabular} & 1    & 2    & 3    & 4    & 5    \\ \midrule
CV                                                              & 0.73 & 1.78 & 3.13 & 4.78 & 6.68 \\ \midrule
V-LSTM                                                          & 0.68 & 1.65 & 2.91 & 4.46 & 6.27 \\ \midrule
C-CGMM+VIM~\cite{cslstm-pre}                                                      & 0.66 & 1.56 & 2.75 & 4.24 & 5.99 \\ \midrule 
S-GAN~\cite{Social_GAN}                                                           & 0.57 & 1.32 & 2.22 & 3.26 & 4.40 \\ \midrule
GAIL-GRU~\cite{gru-compare}                                                        & 0.69 & 1.51 & 2.55 & 3.65 & 4.71 \\ \midrule
CS-LSTM~\cite{cslstm}                                                         & 0.61 & 1.27 & 2.09 & 3.10 & 4.37 \\ \midrule
CS-LSTM (M)~\cite{cslstm}                                                         & 0.62 & 1.29 & 2.13 & 3.20 & 4.52 \\ \midrule
GRIP~\cite{GRIP}                                                            & 0.64 & 1.13 & 1.80 & 2.62 & 3.60 \\ \midrule
PiP~\cite{PiP}                                                             & 0.55 & 1.18 & 1.94 & 2.88 & 4.04 \\ \midrule
HAN~\cite{han}                                                             & \underline{0.42} & \underline{1.01} & \underline{1.69} & \underline{2.52} & \underline{3.54} \\ \midrule 
PECNet[variants]~\cite{pcenet}                                                         & 0.53 & 1.13 & 1.90 & 2.96	& 4.26 \\ \midrule
MID[variants]~\cite{mid}                                                                    & 0.48 & \underline{1.01} & 1.75 & 2.84	& 3.99 \\ \midrule 
Ours                                                            &  \textbf{0.32} &  \textbf{0.83} &  \textbf{1.59} &  \textbf{2.46} &  \textbf{3.52} \\ \bottomrule
\end{tabular}
}
\end{table}

As indicated by the results in Table~\ref{table_comp}, our model achieves state-of-the-art performance.. 
This success can be attributed to its explicit consideration of spatio-temporal features along with intrinsic intentions and uncertainties. 
Following closely is the HAN model~\cite{han}, which demonstrates impressive performance, benefiting from its incorporation of planning information to enrich the complex temporal dynamics of historical data. 
The performance of variants based on PECNet~\cite{pcenet} and MID~\cite{mid} also stands out, representing original implementations inspired by Endpoint-based and Diffusion-based models, respectively. 
These results underscore the significant contributions of endpoint information and diffusion modeling to the advancement of trajectory prediction technology.

\subsubsection{Visualization Analysis}

\begin{figure}[htbp]
  \centering
  \includegraphics[width=1\linewidth]{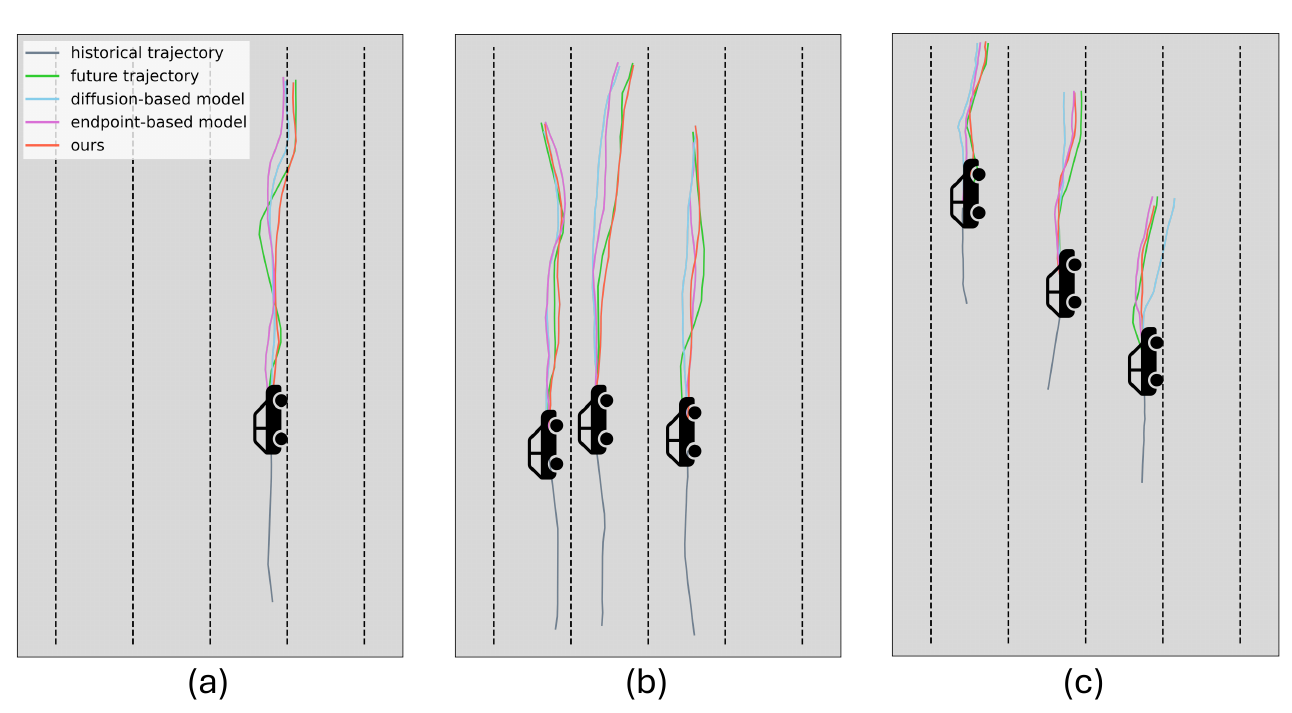}
  \caption{Visualization of road vehicle trajectory prediction.}
  \label{figure-comp}
\end{figure}

The outcomes for road vehicle trajectory prediction across various scenarios are depicted in Figure~\ref{figure-comp}, showcasing our model alongside Endpoint-based and Diffusion-based models.

The integration of endpoint information not only facilitates model convergence but also significantly enhances performance. 
The iterative denoising process inherent to the Diffusion model effectively minimizes the uncertainty associated with future trajectories, thereby yielding accurate predictions. 
Our model leverages the endpoint distribution generated by the Diffusion model to calibrate predicted endpoints. 
By explicitly modeling spatio-temporal features alongside intrinsic intent, it seamlessly amalgamates the strengths of both models, culminating in the most superior performance.

\begin{figure}[htbp]
  \centering
  \includegraphics[width=0.8\linewidth]{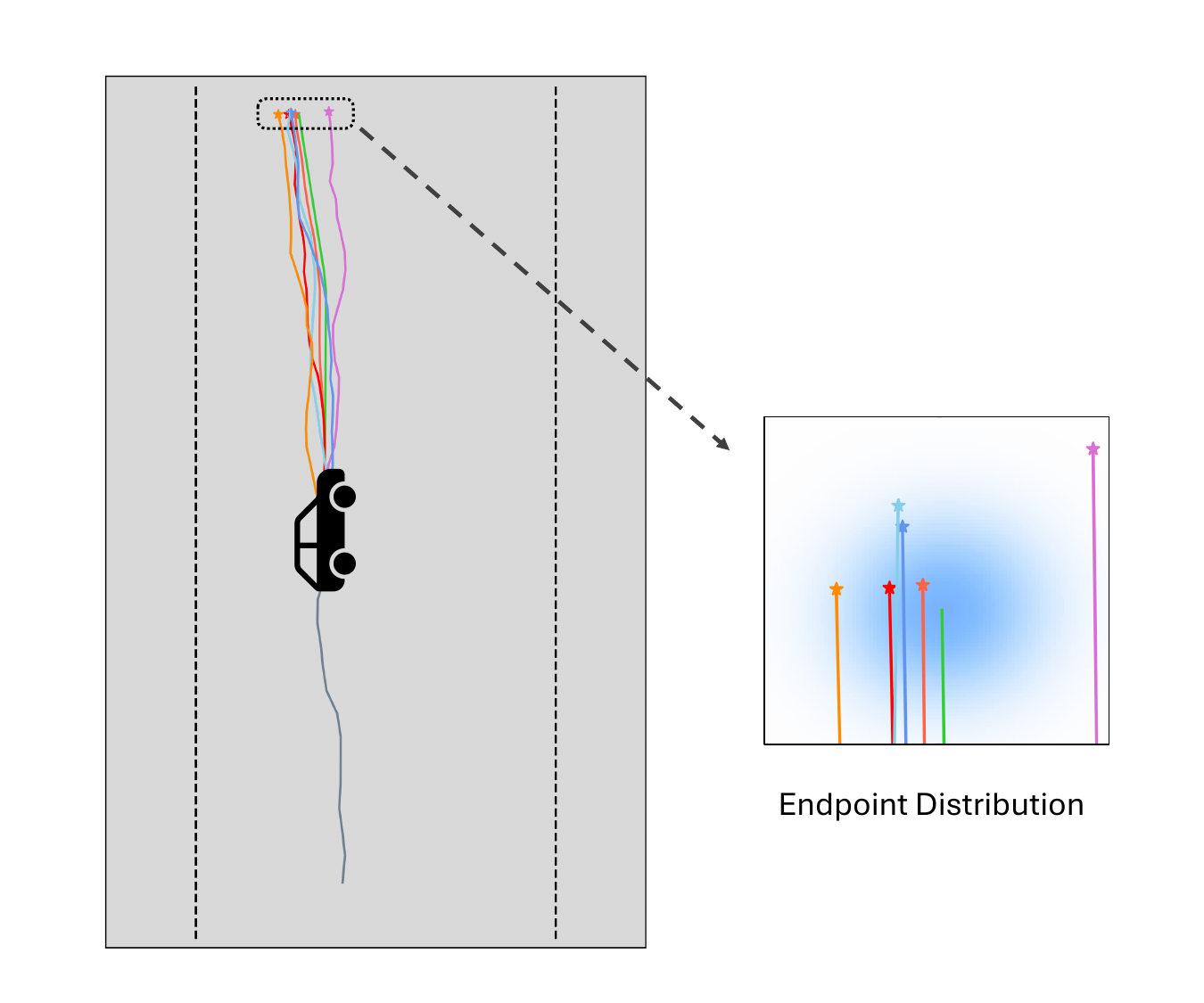}
  \caption{Visualization of endpoint calibration. The figure zooms in to show the endpoint information, which is in a straight line because it is in the last time interval.}
  \label{figure-gau}
\end{figure}

The process of endpoint calibration is illustrated in Figure~\ref{figure-gau}. 
Here, the endpoint distribution module produces distribution data for endpoints, which subsequently informs the calibration of endpoints forecasted by the endpoint prediction module. 
The endpoint that most closely aligns with the endpoint distribution is selected as the final outcome, optimizing the prediction's accuracy.

\subsubsection{Efficiency Analysis}

\begin{table}[htbp]
\caption{Comparison of model complexity. `rel.' indicates a relative value.}
\centering
\label{table_complex}
\resizebox{0.7\linewidth}{!}{
\begin{tabular}{l|cc}

\toprule 
                      & rel. FLOPs & rel. parameters \\ \midrule
ours                  & 1          & 1               \\ \midrule
Endpoint-based model  & 1.07       & 6.12            \\ \midrule
Diffusion-based model & 5.61       & 2.94            \\ \bottomrule
\end{tabular}

}
\end{table}

The complexity comparison of our model with the Endpoint-based and Diffusion-based models is detailed in Table~\ref{table_complex}. 
The Endpoint-based model, employing CVAE for endpoint generation alongside a high-dimensional feature extractor, significantly surpasses our model in terms of the number of parameters. 
Meanwhile, the complexity of the Diffusion-based model is notably enhanced due to its focus on predicting the entire trajectory, not just the endpoints.
Furthermore, its lack of multimodal prediction capabilities and the necessity of repeating the denoising process multiple times also contribute to a higher model cost.

\subsubsection{Ablation Analysis}

\begin{table}[htbp]
\caption{Ablation experiments.}
\centering
\label{table_aba}
\resizebox{0.9\linewidth}{!}{

\begin{tabular}{l|cc|ccccc}
\toprule 
               & \begin{tabular}[c]{@{}c@{}}Endpoint\\ Prediction\end{tabular} & \begin{tabular}[c]{@{}c@{}}Endpoint\\ Distribution\end{tabular} & 1    & 2    & 2    & 3    & 4    \\ \midrule
original model & w/                                             & w/                                               & 0.32 & 0.83 & 1.59 & 2.46 & 3.52 \\ \midrule
variant 1      & w/o                                            & w/                                               & 0.45 & 1.04 & 1.78 & 2.71 & 3.84 \\ \midrule
variant 2      & w/                                             & w/o                                              & 0.51 & 1.07 & 2.06 & 3.04 & 4.33 \\ \midrule
variant 3      & w/o                                            & w/o                                              & 0.59 & 1.30 & 2.23 & 3.25 & 4.47 \\
\bottomrule
\end{tabular}

}
\end{table}

Finally we conduct ablation studies to ascertain the contribution of each module within our framework. 
These studies compare the original model against three variants, as documented in Table~\ref{table_aba}. 
Each variant omits either the endpoint distribution module or the endpoint prediction module.

Variant 1 surpasses Variant 2 in performance, which can be attributed to the endpoint distribution harboring more comprehensive agent spatio-temporal information compared to the information derived from predicted endpoints alone.
Variant 3 yields the least favorable outcomes, a result of it eschewing any endpoint-related information altogether. 
This series of ablation studies vividly illustrates the critical role that both the endpoint distribution and prediction modules play in enhancing the overall accuracy and reliability of our trajectory prediction model.

\section{Conclusion}

In this study, we introduce the Denoised Endpoint Distribution model for multi-agent trajectory prediction, utilizing only trajectory coordinates to predict the movements of all agents in the scene simultaneously. 
By leveraging the Diffusion model, we model the endpoint distribution of agents, and with the Transformer, we capture the historical features of agents. 
This approach allows for an explicit representation of an agent's spatio-temporal features along with its intrinsic intent and uncertainty.

Through extensive experimentation, we demonstrate the superiority of our model. Looking ahead, we expect to explore the use of Plan instead of Endpoint to further enhance our model's performance. This future direction aims to refine our predictive capabilities, potentially offering even more accurate and insightful trajectory predictions for multi-agent systems.

\bibliographystyle{IEEEtran}
\bibliography{IEEEabrv, main}

\end{document}